%% file: Cerutti_ISLPED.tex
\documentclass[sigconf]{acmart}

\AtBeginDocument{%
  \providecommand\BibTeX{{%
    \normalfont B\kern-0.5em{\scshape i\kern-0.25em b}\kern-0.8em\TeX}}}

\copyrightyear{2020}
\acmYear{2020}
\setcopyright{none}\acmConference[ISLPED '20]{ACM/IEEE International Symposium on Low Power Electronics and Design}{August 10--12, 2020}{Boston, MA, USA}
\acmBooktitle{ACM/IEEE International Symposium on Low Power Electronics and Design (ISLPED '20), August 10--12, 2020, Boston, MA, USA}
\acmPrice{15.00}
\acmDOI{10.1145/3370748.3406588}
\acmISBN{978-1-4503-7053-0/20/08}

\usepackage{atbegshi}
\usepackage{xcolor}

\usepackage{graphicx}
\usepackage[nolist]{acronym}
\usepackage{amsmath} 
\usepackage{color,ulem}
\usepackage{booktabs}
\usepackage{lipsum}
\usepackage{soul}

\usepackage{fp}
\usepackage{pgfplots}
\pgfplotsset{compat=1.13}
\usepackage{pgfplotstable}
\usepackage{capt-of}
\definecolor{cieee0}{HTML}{00629b}
\definecolor{cieee1}{RGB}{255,199,44}
\definecolor{cieee2}{RGB}{232,119,34}
\definecolor{cieee3}{RGB}{186,12,47}
\definecolor{cieee4}{RGB}{119, 37, 131}
\definecolor{cieee5}{RGB}{120, 190, 32}
\definecolor{cieee6}{RGB}{0, 132, 61}
\definecolor{cieee7}{RGB}{0,  159,  223}
\newacro{FC}[FC]{fabric control}
\newacro{AI}[AI]{artificial intelligence}
\newacro{DL}[DL]{deep learning}
\newacro{SED}[SED]{sound event detection}
\newacro{KD}[KD]{knowledge distillation}
\newacro{IoT}[IoT]{internet-of-things}
\newacro{IPS}[IPS]{instructions per second}
\newacro{CNN}[CNN]{convolutional neural network}
\newacro{DNN}[DNN]{deep neural network}
\newacro{GRU}[GRU]{gated recurrent unit}
\newacro{RNN}[RNN]{recurrent neural network}
\newacro{MIPS}[MIPS]{Million Instructions Per Second}
\newacro{SIMD}[SIMD]{single instruction multiple data}
\newacro{MCU}[MCU]{Microcontroller}
\newacro{DSP}[DSP]{Digital Signal Processing}
\newacro{PDF}[PDF]{Probability Density Function}
\newacro{MSE}[MSE]{Mean Square Error}
\newacro{SQNR}[SQNR]{Signal to Quantization-Noise Ratio }
\newacro{MOPS}[MOPS]{Million Operation Per Second}
\newacro{MAC}[MAC]{multiply-accumulate}
\newacro{MFCC}[MFCC]{Mel-frequency cepstral coefficients}
\newacro{FFT}[FFT]{Fast Fourier Transform}
\newacro{DCT}[DCT]{Discrete Cosine Transform}
\newacro{BNN}[BNN]{Binary Neural Network}
\newacro{PULP}[PULP]{Parallel Ultra Low Power}
\newacro{HWCE}[HWCE]{Hardware Convolution Engine}
\newacro{DMA}[DMA]{direct memory access}
\newacro{SoC}[SoC]{system-on-chip}
\newacro{GPIO}[GPIO]{general purpose input/output}
\newacro{NN}[NN]{Neural Network}
\newacro{RTC}[RTC]{real-time clock}
\newacro{SDK}[SDK]{software development kit}
\newacro{STFT}[STFT]{short-time Fourier transform}
\newacro{ISA}[ISA]{instruction set architecture}

\newcommand{\calc}[2]{\FPeval{\calcresult}{round(#2,#1)}\calcresult}

\usepackage[binary-units=true]{siunitx}  
\usepackage{textcomp}
\usepackage{textgreek}
\newcommand{\delete}[1]{}
\newcommand{\rev}[1]{#1}

\newcommand{\floor}[1]{\left\lfloor #1 \right\rfloor}
\graphicspath{{./figs/}}

\newcommand{\secref}[1]{Sec.~\ref{#1}}

\newcommand{\tblref}[1]{Tbl.~\ref{#1}}
\newcommand{\figref}[1]{Fig.~\ref{#1}}

\begin{document}

\title{Sound Event Detection with Binary Neural Networks on Tightly Power-Constrained IoT Devices}

\author{Gianmarco Cerutti}
\affiliation{%
  \institution{Fondazione Bruno Kessler}
  \city{Trento}
  \country{Italy}
}
\affiliation{%
  \institution{ETH Zurich}
  \city{Zurich}
  \country{Switzerland}
}
\email{gcerutti@fbk.eu}

\author{Renzo Andri}
\affiliation{%
  \institution{ETH Zurich}
  \city{Zurich}
  \country{Switzerland}
}
\email{andrire@iis.ee.ethz.ch}

\author{Lukas Cavigelli}
\affiliation{%
  \institution{ETH Zurich}
  \city{Zurich}
  \country{Switzerland}
}
\email{cavigelli@iis.ee.ethz.ch}

\author{Elisabetta Farella}
\affiliation{%
  \institution{Fondazione Bruno Kessler}
  \city{Trento}
  \country{Italy}
}
\email{efarella@fbk.eu}

\author{Michele Magno}
\affiliation{%
  \institution{ETH Zurich}
  \city{Zurich}
  \country{Switzerland}
}
\email{magno@iis.ee.ethz.ch}

\author{Luca Benini}
\affiliation{%
  \institution{ETH Zurich}
  \city{Zurich}
  \country{Switzerland}
}
\affiliation{%
  \institution{University of Bologna}
  \city{Bologna}
  \country{Italy}
}
\email{benini@iis.ee.ethz.ch}

\begin{abstract}
Sound event detection (SED) is a hot topic in consumer and smart city applications. Existing approaches based on \acp{DNN} are very effective, but highly demanding in terms of memory, power, and throughput when targeting ultra-low power always-on devices. 

Latency, availability, cost, and privacy requirements are pushing recent IoT systems to process the data on the node, close to the sensor, with a very limited energy supply, and tight constraints on the memory size and processing capabilities precluding to run state-of-the-art DNNs. 

In this paper, we explore the combination of extreme quantization to a small-footprint binary neural network (BNN) with the highly energy-efficient, RISC-V-based (8+1)-core GAP8 microcontroller. 
Starting from an existing CNN for SED whose footprint (815\,kB) exceeds the 512\,kB of memory available on our platform, we retrain the network using binary filters and activations to match these memory constraints.
\rev{(Fully) binary neural networks come with a natural drop in accuracy of 12-18\% on the challenging ImageNet object recognition challenge compared to their equivalent full-precision baselines.} This BNN reaches a 77.9\% accuracy, just 7\% lower than the full-precision version, with 58\,kB (7.2$\times$ less) for the weights and 262\,kB (2.4$\times$ less) memory in total. With our BNN implementation, we reach a peak throughput of 4.6\,GMAC/s and 1.5\,GMAC/s over the full network, including preprocessing with Mel bins, which corresponds to an efficiency of 67.1\,GMAC/s/W and 31.3\,GMAC/s/W, respectively. Compared to the performance of an ARM Cortex-M4 implementation, our system has a 10.3$\times$ faster execution time and a 51.1$\times$ higher energy-efficiency.
\end{abstract}

\keywords{Binary Neural Networks, Sound Event Detection, Ultra Low Power}

\maketitle

\section{Introduction}
\label{sec:intro}
Cloud computing is the most widely-adopted paradigm for deploying \ac{AI} and specifically DNNs to extract useful information from sensors in the \ac{IoT} era \cite{gubbi2013internet}. However, this cloud-centric approach has several drawbacks: high latency due to communication delays, availability and reliability limited by the communication infrastructure, privacy issues due to the streaming of sensitive data to a remote site, and high energy cost for data transmission~\cite{premsankar2018edge}. 
Edge computing is the novel alternative to address these limitations by pushing AI close to the sensors, transmitting only relevant information and alerts \cite{farella2017technologies}. Typically, \ac{IoT} end-nodes are battery-powered and target a long battery life---ideally aiming at self-sustainable operation with the help of energy harvesters, whose collected energy is far from sufficient to power high-performance processors or GPUs \cite{alioto2017iot}.  
\acp{MCU}, with their low power consumption and low cost, are the platform of choice to enable the migration of AI to the edge. 
The leading \ac{MCU} architecture is the ARM Cortex-M series with power consumption in the range of milliwatts and throughput in the order of MOPS. 
To overcome this constrain, over the last few years, many researchers put effort into specialized hardware and optimized inference algorithms to run such DNNs on power-constrained devices. On the software side, network complexity reduction while preserving the quality of predictions is of significant interest in porting deep and complex architectures on a heavily constrained \ac{IoT} node.
There are several approaches to target this goal, e.g., knowledge distillation \cite{hinton2015distilling}, network pruning \cite{he2017channel}, or network quantization \cite{lin2016fixed}. However, only a few implementations of DNNs on microcontrollers are presented in the literature \cite{kusupati2018fastgrnn, zhang2017hello, cerutti2020compact}. An extreme case of quantization is Binary Neural Network (BNN), in which all the weights and activations are described by a single bit representing the value of -1 or 1 \cite{Rastegari2016}. As a consequence, BNNs significantly reduce the amount of memory required and compress 32 MAC operations in just two operations without significantly compromise the accuracy \cite{Rastegari2016}. These two advantages make BNNs a promising approach when resource-constrained devices are involved in edge computing. 

On the hardware side, new approaches enabling near-threshold parallel computing in the MCU space have been explored by researchers, industry, and academia \cite{dreslinski2010near}.
For instance, a novel parallel processor, based on the RISC-V ISA has been launched recently \cite{GapNews2018}. GAP8 is a commercial processor, implemented from the \ac{PULP} open-source project\footnote{https://www.pulp-platform.org}. This processor has similar power requirements of the Cortex-M family (hundreds of mW) with up to 20 times higher computation performance for machine learning applications \cite{GapNews2018}. Furthermore, it features RISC-V extensions providing accelerating the BNN processing. The \texttt{popcount} instruction boosts the processing significantly for BNNs and other quantified neural networks.  

Looking at applications, scene understanding, and context analysis are among the application domains where edge processing can be crucial. They often rely on computer vision. However, the combination with audio processing can highly improve the accuracy of event detection and activity recognition, complementing vision where line-of-sight occlusions or environmental light changes occur \cite{vu2006audio}. Furthermore, the use of audio detection alone can partially solve privacy concerns. 
Thus, \ac{SED} is a powerful tool for many applications such as traffic monitoring \cite{na2015acoustic}, crowd monitoring \cite{meng2015influence}, measurement of occupancy levels for smart and energy-efficient buildings \cite{uziel2013networked}, and emergencies detection \cite{gerosa2007scream}. 

This paper proposes a novel \ac{BNN} for resource constrain and low power microcontrollers for SED applications, i.e. classifying which sound event is present in an audio record.
The proposed BNN has been implemented on the Greenwave's GAP8.  
\newline
\noindent The main contribution of this paper is as follow:
\begin{enumerate}
    \item We propose, train, and efficiently implement a novel \ac{BNN} architecture for \ac{SED}, comparing it with a full-precision baseline network.
    \item We present the design of a full system, based on the low-power and \ac{ISA} optimized for GAP8 microcontroller. The full pipeline is developed from audio acquisition with a low-power microphone, over the Mel bins feature extraction to the on-board classification. We present a detailed analysis of throughput and energy trade-off in a variety of supported configurations as well as on-board measurements.
    \item We demonstrate that binarization of weights and activations are the key factor in matching hardware constraints. Experimental evaluation shows that our implementation on the \ac{PULP} platform is 51x more efficient and 10x faster than the implementation of the same network in the Cortex-M4 based counterpart.
\end{enumerate}

\section{Related Work} \label{sec:relatedwork}

The most used techniques to address \ac{SED} and in general audio processing,  are employing \ac{MFCC} features followed by a GMM, HMM, or SVM classifier \cite{Mesaros2010,Temko-2007,Zhuang20101AED}. Recently, \acp{DNN} \cite{DCASE2017challenge}, \acp{CNN} \cite{hershey2017cnn}, and \acp{RNN} \cite{cerutti2019IS} have been used instead. However, those models require a large amount of memory to perform high-performance predictions: for instance, \acp{DNN} for \ac{SED} such as L3 \cite{cramer2019look} and VGGish \cite{hershey2017cnn} require approximately 4M and 70M parameters, respectively.

Achieving a reduction of the structure size of an existing network for \ac{SED} has been largely investigated in the recent literature. 
In particular, knowledge distillation has been deployed to compress the L3 network to edge-L3 in \cite{cramer2019look}, and VGGish is further compressed to baby VGGish in \cite{cerutti2019IS}.

By replacing the fully connected layer of an existing \ac{CNN} with average max-pooling, Meyer et al. \cite{meyer2017efficient} reduced the number of parameters while increasing the accuracy for the targeted dataset. Still, Meyernet is not suitable for our very constrained IoT use-case. Therefore further model compression is required to match these constraints. 

In addition to model structure modification, recent works on \ac{CNN} have investigated quantization to reduce the storage and computational costs of the inference task \cite{lin2016fixed, lai2018cmsis, iandola2016squeezenet}. As an extreme case of quantization, \acp{BNN} reduce the precision of both weights and neuron activations to a single-bit \cite{Courbariaux2016, Rastegari2016}. \rev{BNNs work on simple tasks like MNIST, CIFAR-10, and SVHN without drop in accuracy \cite{hubara2016binarized}. On the challenging ImageNet dataset, BNNs/TNNs have a drop of 12\%/6.5\% \cite{zhou2016dorefa, spallanzani2019additive}. Recent approaches use multiple binary weight bases, or part of the convolutions are done in full-precision. An accuracy drop down to 3.2\% has been achieved \cite{zhuang2019structured}; unfortunately, these approaches increase the weight memory footprint and computational complexity.}

BNNs are suitable to be implemented on resource-constrained platforms, thanks to their reduced memory requirements and their potential to convert multiplications in hardware-friendly \texttt{XNOR} operations. 

Peak throughput and energy efficiency are achieved by ASIC accelerators. Particularly, BinarEye \cite{moons2018binareye} achieves an energy efficiency of 115\,TMAC/s/W. But these accelerators are not available on the market, and are usually fixed to few network types.

Several works have implemented \acp{CNN} with fixed-point format and operations, in video domain \cite{cerutti2020compact, palossi2018ultra} and in audio domain, where keyword spotting in Cortex-M4 based microcontroller \cite{zhang2017hello}, Cortex-M0+, and Raspberry Pi based platforms \cite{kusupati2018fastgrnn}.

One of the challenges in this field is the development of energy-efficient \ac{NN} firmware implementation for embedded systems. 

Wang et al. \cite{wang2019fannonmcu} developed a library for neural network porting from the FANN framework to ARM \acp{MCU} and \ac{PULP} platforms. In this case, the hardware is fully utilized, but there is support only for multilayer perceptrons. 
Garofalo et al. developed a custom library for quantized convolutional neural networks on PULP \cite{garofalo2019pulp}. However, their focus has been on the precision-throughput trade-off, thereby omitting several optimizations specific to the corner case of binary neural networks and limiting the evaluations to a synthetic single-layer benchmark.

To the best of our knowledge, this is the first BNN proposed and implemented on a parallel RISC-V based microcontroller.

\section{Feature Extraction and BNN} \label{sec:BNN}
The idea behind \acp{BNN} is to approximate the multi-bit filter weights and inputs with binary values in NNs. Binary weights and activations imply a significant decrease in memory usage as well as computational cost \cite{Rastegari2016}.
In this section, we describe the structure of the network, starting from the audio stream to the final prediction.

\subsection{Feature Extraction (Mel Bins)}
\label{sec:FE}
The preprocessing part computes the \ac{STFT} in windows of \SI{32}{\milli\second} every \SI{8}{\milli\second}. Then, we apply the Mel filters to generate 64 Mel bins. The 400 features are then assembled to create the Mel-spectrogram for \SI{3.2}{\second} of audio. The resulting matrix with a shape of $64\times400$ is the input to the neural network. 

\subsection{First Layer and Binarization}\label{subsec:firstLayer}
The input data to the network is non-binary and has, therefore, to be treated separately. A robust approach is to keep the first network layer in full-precision, like in Courbariaux et al. \cite{Courbariaux2016}. In this way, the network learns the binarization function from the training set.

After the convolution, batch normalization is applied, which can be replaced in inference by a bias and a scaling factor, and is finally followed by the signum activation function for binarization. 

To avoid floating-point  operations, all the operations described in this section are done in fixed-point. Fixed-point operations are more efficient in terms of execution time and energy consumption without significant loss of performance \cite{lin2016fixed} also in floating-point embedded systems, and will be evaluated more in detail in the experimental result section.

On the other hand, fixed-point quantization requires additional effort in finding the correct amount of integer and fractional bits for each parameter representation. For doing this, we check the range of the parameters, and we choose the number of integer decimals that represents most of the numbers (99.9\%) without overload error.

\subsection{Binary Convolution}\label{subsec:binaryconvolution}
\ac{BNN}s constrain weights and inputs to $\textbf{I} \in \{-1, 1\}^{n_{in}\times h\times b}$ and $\textbf{W} \in \{-1, 1\}^{n_{out}\times n_{in}\times k_y \times k_x} $. To avoid using two bits, we represent $-1$ with $0$, whereas the actual binary numbers are indicated with a hat (i.e., $\hat{i}=(i+1)/2$). It turns out that multiplications become \texttt{xnor} operations $\bar{\oplus}$ \cite{Rastegari2016}. Formally the output $o_k$ of an output channel $k\in \{0, ..., n_{out}-1\}$  can be described as\footnote{For simplicity, we omit bias and scaling factor in the formula.}:

\resizebox{0.87\linewidth}{!}{
  \begin{minipage}{\linewidth}
  \[
\mathbf{o_k} = \text{sgn}\left(\sum_{n=0}^{n_{in}-1}{{\mathbf{i_n} \ast \mathbf{w_{k,n}}}}\right) = \text{sgn}\left(\sum_{n=0}^{n_{in}-1}{2\left({\mathbf{\hat{i}_n} \ast \mathbf{\hat{w}_{k,n}}}\right)-k_yk_x}\right)\]
\[
= \text{sgn}\left(\sum_{n=0}^{n_{in}-1}{\sum_{(\Delta x, \Delta y)}{2\left({{\hat{i}_n}^{y\text{+}\Delta y,x\text{+}\Delta x} \bar{\oplus} {\hat{w}_{k,n}}}^{\Delta y, \Delta x}\right)-1}}\right)\]
  \end{minipage}
}
\newline\vspace{1mm}\newline
Whereas $\Delta y$ and $\Delta x$ are the relative filter tap positions (e.g., $(\Delta y, \Delta x)\in\{-1,0,1\}^2$ for $3\times 3$ filters). As calculating single-bit operations on microcontroller is not efficient, we pack several input channels into a 32-bit integer (e.g., the feature map pixels at $(y+\Delta y,x+\Delta x)$ in spatial dimension and input channels $32n$ to $\left(32(n+1)-1\right)$ packed in $\mathbf{\hat{i}_{32n:+32}}^{y\text{+}\Delta y,x\text{+}\Delta x}$), while the Multiply Accumulates (MACs) can be implemented with \textit{popcount} and \textit{xnor} operations. 

Furthermore, as common embedded platforms like GAP8 do not have a built-in \textit{xnor} operator, the \textit{xor} operator $\oplus$ is used and the result is inverted. Therefore, the final equation for the output channel is $\mathbf{o_k}=$\newline 
\resizebox{1\linewidth}{!}{
  \begin{minipage}{\linewidth}
  \begin{align*}
\text{sgn}\left(\sum_{n=0}^{\frac{n_{in}}{32}-1}{\sum_{(\Delta x, \Delta y)}{32-2\text{popcnt}\left(\mathbf{\hat{i}_{32n:+32}}^{y\text{+}\Delta y,x\text{+}\Delta x} {\oplus} \mathbf{\hat{w}_{k,32n:+32}}^{\Delta y, \Delta x}\right)}}\right)
\end{align*}
  \end{minipage}
}
\newline\vspace{-4mm}\newline

\subsection{Batch Normalization and Binarization}
A batch normalization layer follows each binary convolutional layer. As the output of binary layers are integer values, and the signum function can be written as a comparison function, the activation function is simplified to:
\begin{equation}
    \text{binAct}(x) = \begin{cases} 0, & \mbox{if } x \cdot \text{sgn}(\gamma') \geq \floor{\frac{\beta'}{\gamma'}} \\[1mm] 1, & \mbox{if } x \cdot \text{sgn}(\gamma') < \left\lfloor \frac{\beta'}{\gamma'} \right\rfloor \end{cases}.
\end{equation}
whereas $\gamma'$ is the scaling factor and $\beta'$ is the bias based on the batch normalization parameters. While exporting the model, we compute the integer threshold value $\lfloor\frac{\beta'}{\gamma'}\rfloor$ in advance.
In inference, one sign comparison and one threshold comparison have to be calculated for each activation value.

\subsection{Last Layer and Prediction}
In the last layer, the fixed-point values from the last binary layer are convolved with the fixed-point weights, and N output channels are calculated, where N is the number of classes. Finally, the network performs an average pooling over the whole image giving N predictions for each class.
\subsection{Neural Network Architecture}
\label{sec:neural_network_architecture}
\tblref{tbl:network_summary} summarizes the architecture of the NN. The neural network consists of 7 hidden layers, 5 of which are binary. The first and last layers are real-valued. 
Their required computations are significantly smaller than in the binary layers (e.g., 7\,MMAC in the first layer compared to 109\,MMAC in the second layer), and therefore they minimally contribute to the overall computational effort. 
The reason for having real-valued layers is the high loss of accuracy with entirely binarized neural networks \cite{Rastegari2016}.

\begin{table}
    \centering
    \caption{Kernel size, channel, and computational effort for each layer.}
    \label{tbl:network_summary}
    \begin{tabular}{@{}lrrrr@{}}
    \toprule
    \textbf{Layer} & \textbf{Kernel Size} & \textbf{Channel} & \textbf{Stride} & \textbf{MACs}\\
    \midrule
    First (real-valued) & 3 $\times$ 3 & 32 & 1 & 7M\\
    1. Binary Layer & 3 $\times$ 3 & 64 & 2 & 109M\\
    2. Binary Layer & 3 $\times$ 3 & 128 & 1 & 405M\\
    3. Binary Layer & 3 $\times$ 3 & 128 & 2 & 186M\\
    4. Binary Layer & 3 $\times$ 3 & 128 & 1 & 154M\\
    5. Binary Layer & 1 $\times$ 1 & 128 & 1 & 17M\\
    Last (real-valued) & 1 $\times$ 1 & 28 & 1 & 6M\\
    \midrule
    \textbf{Total:} & & & & 884M\\
    \bottomrule
    \end{tabular}
    
\end{table}

\begin{figure}[ht]
\centering
\includegraphics[width=0.9\columnwidth]{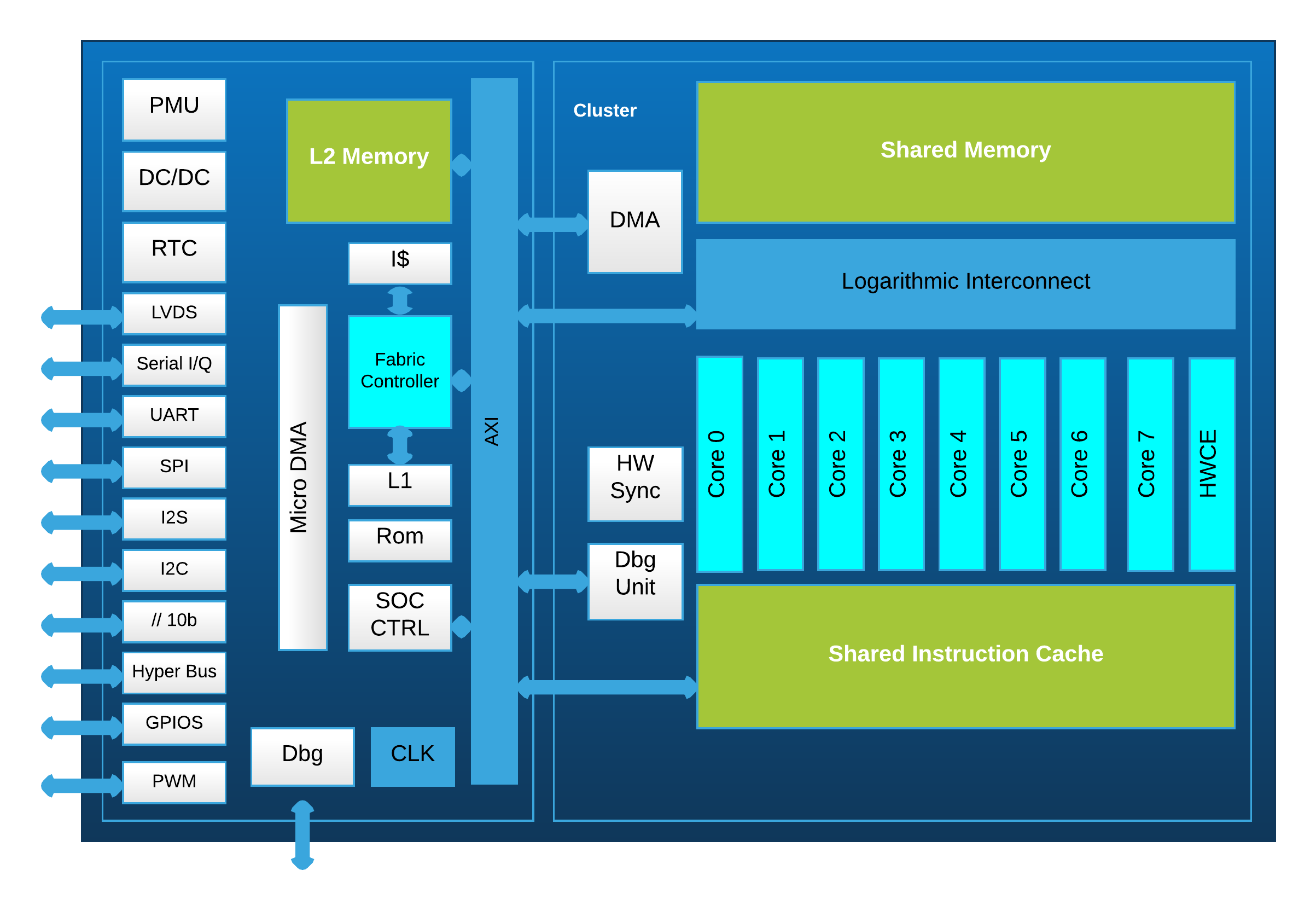}
\vspace{-5mm}
\caption{Architecture of GAP8 embedded processor \cite{flamand2018gap}}
\label{fig:gap_arch}
\vspace{-1cm}
\end{figure}

\section{Embedded Implementation} \label{sec:GAP8}
The Mel bins extraction and BNN are implemented on GAP8. 
The application scenario for this device is low-latency low-power signal processing. The device has a tunable frequency and voltage supply.
\figref{fig:gap_arch} shows the main block of the chip: GAP8 has two main programmable components, the \ac{FC}, and the cluster. The \ac{FC} is the central microcontroller unit, and it is meant to manage peripherals and offload workloads to the cluster. The cluster is composed of eight parallel RISC-V cores, a convolution accelerator, and shared memory banks. The two domains share the same voltage source but keep two different frequencies: On-chip DC-DC converters translate the voltage, and two independent frequency-locked loops (FLLs) generate the two different clock domains. The \ac{FC} is a single-core in-order microcontroller implementing the RISC-V instruction set. To customize the core for signal processing application, GAP8 extends the RISCV-IMC instruction set for signal processing application. In addition to integer, multiplication, and compressed instruction (IMC), GAP8 ISA supports Multiply and Accumulate, Single Instruction Multiple Data (SIMD), Bit manipulation, post-increment load/store, and Hardware Loops. The \ac{FC} is directly interconnected to an L2 memory of \SI{512}{\kilo\byte} SRAM.
The cluster has eight cores identical to the \ac{FC}. The cores share the 64\,kB L1 SRAM scratchpad memory, equipped with a logarithmic interconnect that supports single-cycle concurrent access from different cores requesting memory locations on separate banks.

The cores fetch instructions from a multi-ported instruction cache to maximize the energy efficiency on the data-parallel code. \rev{Moreover, an efficient DMA (called \textmu DMA) enables multiple direct transfers from peripherals and L1 to the L2 memory.} The cluster has a hardware synchronizer for event management and efficient parallel threads dispatching. The \ac{FC} and cluster communicate with each other by an AXI-64 bidirectional bus. The software running on the \ac{FC} overviews all tasks offloaded to the cluster and the \textmu DMA. At the same time, a low-overhead runtime on the cluster cores exploits the hardware synchronizer to implement shared-memory parallelism in the fashion of OpenMP \cite{conti2016enabling}.

\section{Experimental Results} \label{sec:experiments}
To accurately evaluate the BNN, we designed a full system. Thus, the power and energy-efficient measurements are performed on the hardware platform. \delete{For training and evaluating the algorithm, we used an open dataset that we illustrate in the next subsection.} 

\subsection{Dataset} 
In this work, we use the dataset of Takahashi et al. \cite{Takahashi2016DeepRecognition}, which is based on the Freesound database, an online collaborative sound database. It consists of 28 different event types, e.g., instruments, animals, mechanical sounds. Each clip has a variable length, and the total length of all 5223 audio files is 768 minutes. All audio samples have a sampling rate of \SI{16}{\kilo\hertz}, a bit depth of 16, and are single-channel. The dataset is split into training (75\%) and test set (25\%).
We compute the \ac{STFT} in windows of 512 samples every 128 samples, respectively \SI{32}{\milli\second} and \SI{8}{\milli\second}. Then we apply 64 Mel-filters to generate 64 Mel bins. 400 features are then tiled together to create the Mel-spectrogram for \SI{3.2}{\second} of audio (see \secref{sec:FE}). For the training set, we split each audio clip in consecutive chunks of \SI{3.2}{\second}.

\delete{If a chunk is shorter than 3.2 seconds, it is discarded, but if the audio clip is shorter than 3.2\,s, the only one chunk is padded with zeroes, so that each audio clip has at least one chunk associated.}\rev{Chunks shorter than 3.2s are discarded, or zero-padded if it is the only chunk.} In the test set, we extract one single patch of 3.2\,s, starting from half of the clip.

\subsection{Firmware Details} \label{sec:firmware}
To cope with L1 memory constraints, we run the prediction on 4 tiles in which the image is split. The tiles have an overlap of 20 pixels to take into account the receptive field of convolutional kernels at the border of the tiles. 
The firmware implements a double buffering for the weight loading: before the program processes the input of a specific layer, the cores configure the DMA to load the weights of the next layer, from the L2 memory to the single-cycle accessible L1 memory. 
An interesting feature of GAP8 is the built-in \texttt{popcount} instruction, which takes just one cycle and decreases the execution time significantly in binary layers, thus useful for BNN calculation. The single 3$\times$3$\times$C kernel application gains speed thanks to loop unrolling. Finally, the code parallelization over the eight cores is implemented using the OpenMP API.

\subsection{Accuracy}
We start from MeyerNet \cite{meyer2017efficient} and use the Additive Noise Annealing (ANA) algorithm \cite{spallanzani2019additive} to train the network with binary weights and activations. \tblref{tbl:accuracy} provides an overview of the original MeyerNet and the BNN. 
The BNN-GAP8 network keeps the first and the last layer in 16-bit fixed-point, whereas the other layers are binary. 
For the accuracy of Meyernet, we consider its 16-bit quantized version because it is expected\footnote{DNNs are robust to quantization down to 16\,bit \cite{lai2018cmsis, palossi2018ultra, jacob2018quantization}} to be the same the FP32 baseline. 

\begin{table}
\centering
\caption{Accuracy and Memory Footprint for the Baseline CNN (16-bit Fixed-Point precision), BNN with first/last layer in 16-bit Fixed-Point.}\label{tbl:accuracy}
\begin{minipage}{1\columnwidth}
\centering
\begin{tabular}{@{}lrr@{}}
\toprule
         & CNN \cite{meyer2017efficient}& BNN-GAP8 \\ \midrule
Accuracy & 85.1\% & \textbf{77.9\%}\\ 
Memory for weights [kB] & 815 & 58 \\
Memory for input [kB] & 204 & 204 \\
Memory requirement [kB] & 1019\footnote{It does not fit into the 512\,kB SRAM of the GAP8 microcontroller.} & 262 \\\bottomrule
\end{tabular}
\end{minipage}
\end{table}

The BNN achieves an accuracy of 77.9\%, which is 7.3\% below the full-precision baseline and is in-line with state-of-the-art \rev{binary and ternary networks (i.e., 12\% binary and 6.5\% ternary neural networks for ImageNet  \cite{zhou2016dorefa, spallanzani2019additive}).}

\tblref{tbl:accuracy} shows that the \ac{BNN} matches with the memory constraints of \SI{512}{\kilo\byte} of L2 memory in GAP8 chip, in contrast to the fixed-point baseline.

\subsection{Energy Efficiency}\label{sec:sedenergyefficient}
In the following section, we are discussing the throughput and energy efficiency trade-off. First, we sweep the independent cluster and fabric control frequency $(f_{cl}, f_{fc})\in \{30,50,85,100,150\}\,\text{MHz}\ \times$ $\{10,30,50,100,150\}\,\text{MHz}$ for \SI{1}{\volt}, and $(f_{cl}, f_{fc})\in$ $\{50,100,150,200,\\250\}\,\text{MHz}\times\{10,30,50,100,150\}\,\text{MHz}$ for \SI{1.2}{\volt}, supported by the GAP8 microcontroller.
We set the real-time constraint to 0.3125 frames per second due to the 3.2\,s long audio samples.

\begin{figure}
\centering
\includegraphics[width=1\columnwidth]{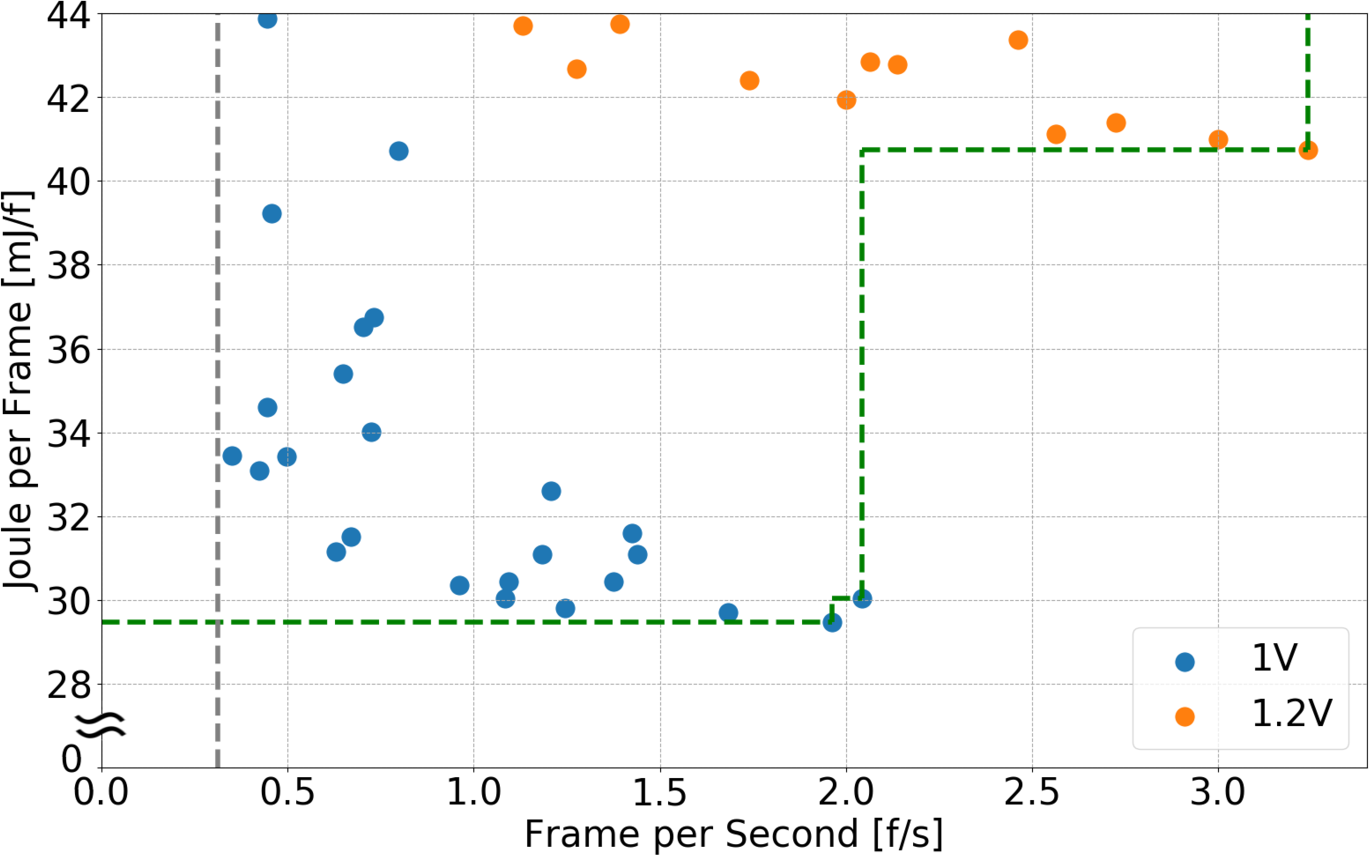} 
\caption{Throughput and energy efficiency at different supply voltages and operating frequencies. All of the measured settings fulfill the requirement of one classification every 3.2s (see the grey dashed line).}\label{fig:paretocurve}
\end{figure}

\figref{fig:paretocurve} shows clearly that the 1.0\,V corners pareto-dominate the faster 1.2\,V corners. It can be seen that the most energy-efficient corner is at 100\,MHz for the \ac{FC}, and 150\,MHz for the cluster, where the system achieves an energy efficiency of 31.3 GMAC/s/W, and a throughput of 1.5 GMAC/s.

\subsection{Execution Time and Power Consumption}
We profile time and throughput as well as the energy-efficiency of each layer of the \ac{NN}. The network architecture is shown in \tblref{tbl:network_summary} together with the amount of \ac{MAC} required for each layer at the most energy-efficient corner according to the analysis in the previous section (i.e., $V_{dd}=1.0\,\text{V}$, $(f_{cl}, f_{fc})=(150\,\text{MHz}, 100\,\text{MHz})$).

The measurements are performed with the \textit{Rocketlogger} \cite{SGLLLT2017}. Voltage and current of the \ac{SoC} are logged. We evaluate the power and duration of measurements and calculate the energy consumption. The results for each layer are listed in \tblref{tbl:performance}.

\begin{table}
        \caption{Duration and energy consumption for each layer as well as throughput and energy efficiency compared to MACs.}
    \label{tbl:performance}\label{tab:breakout}
    \scalebox{0.89}{
    \begin{tabular}{@{}lrrrrr@{}}
    \toprule
         \textbf{Layers} & \multicolumn{1}{c}{\textbf{MACs}} & \multicolumn{1}{c}{\textbf{Time}} & \multicolumn{1}{c}{\textbf{Energy}} & \multicolumn{1}{c}{\textbf{Through.}} & \multicolumn{1}{c}{\textbf{Efficiency}} \\
         & &[ms]&[mJ]& \multicolumn{1}{c}{[MAC/s]} & \multicolumn{1}{c}{[MAC/s/W]}\\
    \midrule
         Mel bins & - & 77.0 & 2.64 & - & -\\
         First Layer     & 7M & 130.8 & 5.94 & 54M & 1.2G\\
         1. Bin Layer & 109M & 73.3 & 3.57 & 1494M  & 30.6G\\
         2. Bin Layer & 404M & 168.0 & 8.86 & 2404M & 45.6G\\
         3. Bin Layer & 185M & 51.2 & 2.94 & 3628M & 63.2G\\
         4. Bin Layer & 154M & 40.3 & 2.29 & 3822M & 67.1G\\
         5./6. Layer\footnote{The two last layers are merged in the implementation.} & 21M  & 47.4 & 1.93 & 1724M  & 1.9G\\
    \midrule
        \textbf{Total/Average} & 882M & 588.0 & 28.18 & 1503M & 31.3G\\
        \bottomrule
    \end{tabular}
    }%

\end{table} 
\def\brtotalheight{140}
\def\brymaxFIRST{22} 
\def\brymaxSECOND{88} 
\def\brfullwidth{1.25\linewidth}
\def\brlegendwidth{1.07\linewidth}
\def\brlegendspacing{0.1}
\newcommand{\rnnfigures}[1]{#1}

\begin{figure}
\input{./figs/comparison_plot.tex}
\vspace{-0.5cm}
\caption{Improvement in throughput and energy efficiency compared to the ARM Cortex-M4 implementation.}\label{fig:bnngap_comparison}
\end{figure}
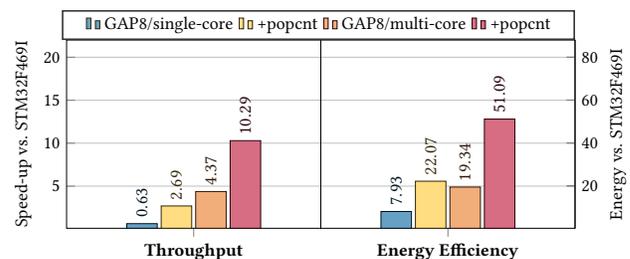

Binary layers are the most efficient ones; this is because of the combination of \texttt{xor} and \texttt{popcount} instructions process\rev{ing} 32 pixels in just 2 instructions. The efficiency peak is at 67.1\,GMAC/s/W in the fourth binary layer, and the average efficiency is 34.5\,GMAC/s/W. The most efficient configuration meets the real-time constraint, and the entire network runs within \SI{0.511}{\second}.

For a further investigation of the improvement in throughput and energy efficiency thanks to the capabilities of the GAP8 SoC, we have implemented the BNN on the STM32F469I Discovery board. \figref{fig:bnngap_comparison}  gives an overview of the improvements of the GAP8 implementations compared to the single-core ARM Cortex-M4F implementation, which has \texttt{popcount} implemented in software. We port the SW-\texttt{popcount} (i.e., 12 cycles) to GAP8 and run the code on a single core, and all 8 cores.
The GAP8 compared to the STM32F469I, running both the BNN on a single-core and without HW-\texttt{popcount}, shows a 7.9$\times$ better energy efficiency, but with a \calc{1}{1/0.63}$\times$ lower throughput due to the higher operating frequency of the ARM core. Enabling the HW-\texttt{popcount} gives a significant improvement in energy efficiency (\calc{1}{22.07/7.93}$\times$) and speed in computation (\calc{1}{2.69/0.63}$\times$). Running the BNN on all 8 cores gives an improvement of \calc{1}{4.37/0.63}/\calc{1}{19.34/7.93}$\times$ in throughput and energy efficiency. Finally, the \texttt{popcount} ISA extension gives another boost of \calc{1}{10.29/4.37}$\times$ and \calc{1}{51.09/19.34}$\times$, respectively.

Overall the GAP8 implementation that uses all the functionality of the core (i.e., \texttt{popcount} instruction and multi-core) is 10$\times$ faster and 51$\times$ more efficient than running the same network on the Cortex-M4F.

\ifdefined\REMOVEPOWERTRACE \else
\figref{fig:power_traces} shows the power trace of the layers in the same setup in \tblref{tab:breakout}. As described in \secref{sec:firmware}, we split the input data into tiles to match the memory constraints. The traces refer to one tile out of four. Thus the execution time is approximately one-fourth of the one presented in \tblref{tab:breakout}.
Between layers, the \ac{FC} offloads the cluster for configuring the next layer: it switches the input and output buffer, allocates memory for the next weights, configures the DMA, and so on. This behavior is visible in the drop of power traces because the cluster is in sleep, and the activity of the \ac{FC} consumes less. Similar behavior can be observed inside binary layers, where the processing is split in chunks of 32 channels.

\begin{figure}
\centering
\includegraphics[width=1\columnwidth]{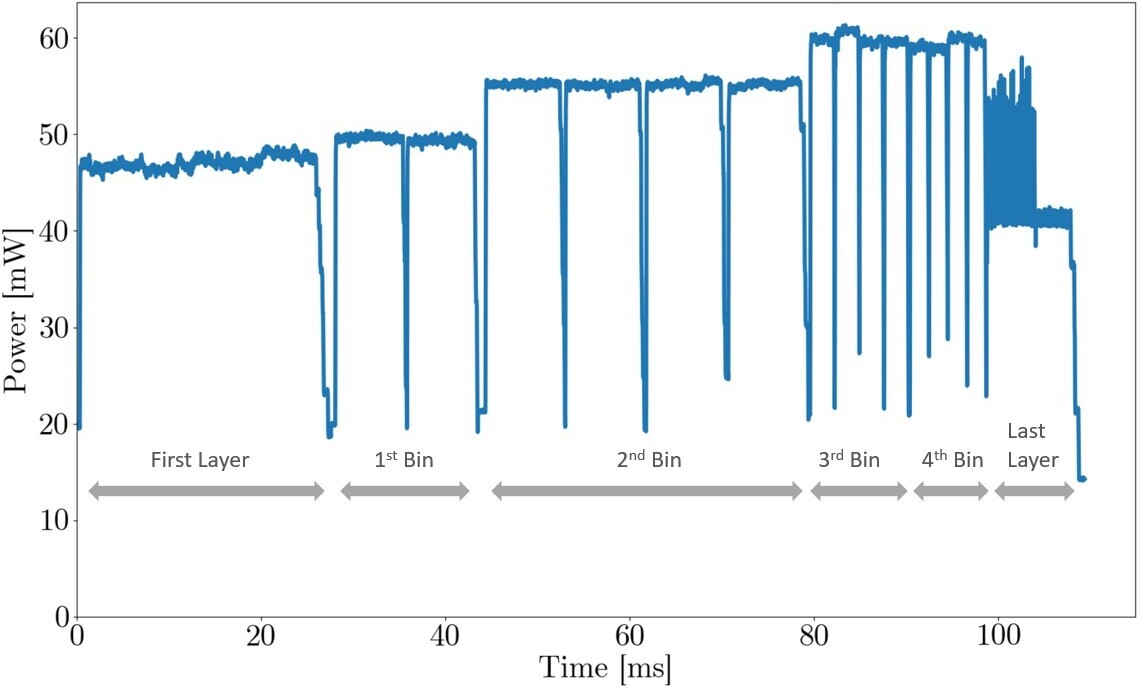}
\caption{Power trace of running the BNN on one tile on the GAP8 platform.} \label{fig:power_traces}
\vspace{-0.7cm}
\end{figure}\fi

\section{Conclusions} \label{sec:conclusions}
Starting from the best-performing DNN for sound event detection on our target dataset, we have proposed and trained a DNN with the same topology but binary weights and activations. The proposed BNN matches the memory and resource-constraints of milliwatt range of the target embedded platforms. The resulting BNN has an accuracy of 77.9\%, a drop of 7.2 percent point from the full-precision baseline \rev{which is in line of similar state-of-the-art BNNs/TNNs (i.e., 6.5-19\%)}. The overall program requires 230\,kB of RAM, \calc{1}{1019/262}$\times$ less than the system using 16-bit quantized baseline CNN. Due to this compression, the network fits in the GAP8 \ac{PULP} Platform. We evaluated energy efficiency with experimental measurement of the power consumption of the full system. The classification of \SI{3.2}{\second} of audio requires \SI{511}{\milli\second} and \SI{25.54}{\milli\joule}, with a peak energy efficiency of 67.1\,GMAC/s/W and average 34.5\,GMAC/s/W. The performance on the GAP8 board has been shown to be 10$\times$ faster and 51$\times$ more energy-efficient than on an ARM Cortex-M4F platform, which comes from multi-core capabilities (i.e., 4.3/19.3$\times$), the build-in \texttt{popcount} instruction (i.e., \calc{1}{10.29/4.37}/\calc{1}{51.1/19.34}$\times$). 

\begin{acks}
This work was in part funded by the U.S. Office of Naval Research Global under the project ONRG - NICOP - N62909-19-1-2018, ``Zero-power sensing for underwater monitoring.''
\end{acks}

\bibliographystyle{ACM-Reference-Format}
\bibliography{refsshortened}

\end{document}

%% file: figs/comparison_plot.tex
\begin{tikzpicture}[scale=0.75]
 
\def\brplotshiftvertical{6.5}
\def\brymax{22} 

\ifdefined\THESIS
\def\brplotshiftvertical{6.5} 
\def\brtotalheight{220}
\def\brymaxFIRST{18} 
\def\brymaxSECOND{72} 
\def\brfullwidth{1.25\textwidth}
\else

\fi

\begin{axis}[ legend style={draw=none},clip=false, x tick label style={ /pgf/number format/1000 sep=}, ymax=\brymaxFIRST,xtick={0,1}, extra y tick style={grid=major}, extra y tick labels={1},  ylabel={Speed-up vs. STM32F469I},enlarge x limits=0, xmax=1.5, enlarge y limits=0, ymin=0.1, ybar, xmin=-0.5,xminorgrids=false, ymajorgrids=true, yminorgrids=true, yminorticks=true, xminorticks=false, 
legend style={anchor=south east,legend columns=-1},nodes near coords,
        xticklabels = {
\textbf{Throughput}, \textbf{Energy Efficiency}
        },
    bar width=0.12, nodes near coords={\pgfmathprintnumber[precision=2]{\pgfplotspointmeta}}, every node near coord/.append style={rotate=90, anchor=west},width=\brfullwidth, height=\brtotalheight
        ] 
%
\addplot[cieee0!20!black,fill=cieee0!60!white] table[x index=0,y index=3, restrict expr to domain={\thisrow{id}}{0:0}] {./figs/comparison_table.csv};
\addplot[cieee1!20!black,fill=cieee1!60!white] table[x index=0,y index=2, restrict expr to domain={\thisrow{id}}{0:0}] {./figs/comparison_table.csv};
\addplot[cieee2!20!black,fill=cieee2!60!white] table[x index=0,y index=4, restrict expr to domain={\thisrow{id}}{0:0}] {./figs/comparison_table.csv};
\addplot[cieee3!20!black,fill=cieee3!60!white] table[x index=0,y index=1, restrict expr to domain={\thisrow{id}}{0:0}] {./figs/comparison_table.csv};

\def\barplotleft{-1.25}
\def\barplotwidth{0.25}
\def\barplotshift{0.30}

\def\brplotleft{0.5}
\def\brplotshift{1.0}
\pgfplotsinvokeforeach{0.0}{\draw (\brplotleft+#1*\brplotshift, 0) -- (\brplotleft+#1*\brplotshift,\brymaxFIRST);};

\ifdefined\THESIS
\def\brlegendwidth{1.10\textwidth}
\def\brlegendspacing{0.1}
\else
\fi
\def\legendBLUE{ GAP8/single-core}
\def\legendYELLOW{ +popcnt}
\def\legendORANGE{ GAP8/multi-core}
\def\legendRED{ +popcnt}
\ifdefined\THESIS
\node at (axis cs:1.5,\brymaxFIRST) [anchor=south east, text centered, draw=black!80, minimum width=\brlegendwidth, fill=white!20!white, line width=0.8pt] {\textbf{Legend: }\includegraphics{\rnnfigures{blue}}\legendBLUE\hspace{\brlegendspacing cm}\includegraphics{\rnnfigures{yellow}}{\legendYELLOW}\hspace{\brlegendspacing cm}\includegraphics{\rnnfigures{orange}}{\legendORANGE}\hspace{\brlegendspacing cm}\includegraphics{\rnnfigures{pink}}{\legendRED}};
\else
\node at (axis cs:1.5,\brymaxFIRST) [anchor=south east, text centered, draw=black!80, minimum width=\brlegendwidth, fill=white!20!white, line width=0.8pt] {\includegraphics{\rnnfigures{blue}}\legendBLUE\hspace{\brlegendspacing cm}\includegraphics{\rnnfigures{yellow}}{\legendYELLOW}\hspace{\brlegendspacing cm}\includegraphics{\rnnfigures{orange}}{\legendORANGE}\hspace{\brlegendspacing cm}\includegraphics{\rnnfigures{pink}}{\legendRED}};

\fi

\end{axis}

\begin{axis}[ legend style={draw=none},clip=false, x tick label style={ /pgf/number format/1000 sep=}, ymax=\brymaxSECOND,xtick={0,1}, extra y tick style={grid=major}, extra y tick labels={1},  ylabel={Energy vs. STM32F469I},enlarge x limits=0, xmax=1.5, enlarge y limits=0, ymin=0.1, ybar, xmin=-0.5,xminorgrids=false, ymajorgrids=true, yminorgrids=true, yminorticks=true, xminorticks=false, 
legend style={anchor=south east,legend columns=-1},nodes near coords,
    bar width=0.12, nodes near coords={\pgfmathprintnumber[precision=2]{\pgfplotspointmeta}}, every node near coord/.append style={rotate=90, anchor=west},width=\brfullwidth, height=\brtotalheight, 
    axis y line*=right, axis x line=none, grid=none
        ] 
    
    \addplot[cieee0!20!black,fill=cieee0!60!white] table[x index=0,y index=3, restrict expr to domain={\thisrow{id}}{1:1}] {./figs/comparison_table.csv};
    \addplot[cieee1!20!black,fill=cieee1!60!white] table[x index=0,y index=2, restrict expr to domain={\thisrow{id}}{1:1}] {./figs/comparison_table.csv};
    \addplot[cieee2!20!black,fill=cieee2!60!white] table[x index=0,y index=4, restrict expr to domain={\thisrow{id}}{1:1}] {./figs/comparison_table.csv};
    \addplot[cieee3!20!black,fill=cieee3!60!white] table[x index=0,y index=1, restrict expr to domain={\thisrow{id}}{1:1}] {./figs/comparison_table.csv};
\end{axis}
\end{tikzpicture}